\pdfoutput=1
\documentclass[runningheads]{llncs}

\usepackage{graphicx}
\usepackage{amsmath}
\usepackage{amssymb}

\usepackage{hyperref}
\usepackage{caption}
\usepackage{subcaption}

\begin{document}

\title{Transfer Learning as an Essential Tool for Digital Twins in Renewable Energy Systems}

\titlerunning{Transfer Learning for Renewable Energy Systems}
\author{Chandana Priya Nivarthi}
\authorrunning{C. Nivarthi}
\institute{Intelligent Embedded Systems Lab, University of Kassel, Germany 
\email{chandana.nivarthi@uni-kassel.de}}

\maketitle              % typeset the header of the contribution

\begin{abstract}

Transfer learning (TL), the next frontier in machine learning (ML), has gained much popularity in recent years, due to the various challenges faced in ML, like the requirement of vast amounts of training data, expensive and time-consuming labelling processes for data samples, and long training duration for models. TL is useful in tackling these problems, as it focuses on transferring knowledge from previously solved tasks to new tasks. Digital twins and other intelligent systems need to utilise TL to use the previously gained knowledge and solve new tasks in a more self-reliant way, and to incrementally increase their knowledge base. Therefore, in this article, the critical challenges in power forecasting and anomaly detection in the context of renewable energy systems are identified, and a potential TL framework to meet these challenges is proposed. This article also proposes a feature embedding approach to handle the missing sensors data. The proposed TL methods help to make a system more autonomous in the context of organic computing.

\end{abstract}

\section{Introduction}
\label{sec:intro}

In recent years, machine learning (ML) has achieved great success in many practical applications. Nonetheless, it still faces limitations for certain real-world scenarios. The primary challenges include lack of availability of sufficient labeled instances for training the models and the fact that future unseen instances of test data might be drastically different from train data. However, collecting and labeling sufficient training data is often expensive, time-consuming, and even unrealistic. When the training and test data have different data distributions, the traditional ML approach shows a degradation in the performance of the model \cite{shimodaira2000}. 

Transfer learning (TL) can help resolve these challenges inherently, and hence it is considered as ML's next frontier \cite{ruder2017} by the research community. TL aims to utilize the existing knowledge of ML models and transfer it to new models where sufficient training data is not available.

TL can be considered analogous to a human's approach when doing tasks. Just as we apply knowledge gained from one task to do another related task, similarly an intelligent system needs to use TL to utilize the knowledge from one task while solving another. This has motivated extensive research in TL and its applications in various domains.

TL is also gaining popularity in the fields of Computer Vision \cite{weiss2016} and Natural Language Processing (NLP) \cite{sogard2013}, where the data are primarily images and text, respectively. In these domains, TL is mostly applied on ML classification problems, like image classification or sentiment classification of texts. In existing literature, TL methods are not often applied to regression problems, especially in the field of renewable energy systems.

Digital Twins (DT) for renewable energy systems like a wind power plant or a solar power plant create an evolving profile of the plant and record data throughout the life-cycle of plant, which helps to give insights on the operations, maintenance and configurations of the plant. When equipped with ML-based predictive capability, these DT substantially increase the efficiency and accuracy with which real-time decisions are taken. 

ML is primarily useful in predicting the day-ahead power generation or detecting anomalies in power plants, both of which require historical data. But often when a new plant is installed, we do not have training data available, and the internal and external influences on the power plant change over time. However, one would like to predict the power generated by that new plant as quickly as possible and to consider the changing conditions. Here TL is essential, as we can transfer the knowledge gained from similar existing power plants to the newly set up ones.

The remainder of this article is structured as follows. Section \ref{sec:related} gives a general overview of TL terminology and summarizes the related work and literature related to this field. Section \ref{sec:proposal} details this article's research direction towards TL framework and applications of TL in relevant use cases. Section \ref{sec:progress} summarizes the current progress in our research and gives an outlook on future work.

\section{Related Work}
\label{sec:related}
In this section, TL notations, definitions, and a general overview of TL are discussed in subsection \ref{subsec:tl_notations}. Then, we introduce existing literature on TL applications in renewable energy systems in subsection \ref{subsec:tl_def}.

\subsection{Notations and Definitions in Transfer Learning}
\label{subsec:tl_notations}
In this subsection, we will briefly discuss the notations used in the TL research community \cite{ruder2017,pan2010}.  Before defining TL, we define the terms \textit{domain} and \textit{task} in the context of TL.

A domain $\mathcal{D} = \{\mathcal{X}, P(X)\}$ is defined by two components: a feature space $\mathcal{X}$ and a marginal probability distribution $P(X)$, where $X \in \mathcal{X}$. If the source domain, $\mathcal{D}_s$ and target domain, $\mathcal{D}_t$ are different, then they either have different feature spaces ($\mathcal{X}_s \neq \mathcal{X}_t$) or different marginal distributions ($P(X_s) \neq P(X_t)$).

Given a specific domain $\mathcal{D}$, a task $\mathcal{T}$ consists of a label space $\mathcal{Y}$ and a conditional probability distribution $P(Y|X)$ and is represented as $\mathcal{T} = \{\mathcal{Y},P(Y|X)\}$, or $\mathcal{T} = \{\mathcal{Y},\mathbf{f}(\cdot)\}$, where $\mathbf{f}(\cdot)$ is learned from the training data consisting of input $X \in \mathcal{X}$ and output $Y \in \mathcal{Y}$. If the source task, $\mathcal{T}_s$ and target task, $\mathcal{T}_t$ are different, then they either have different label spaces ($\mathcal{Y}_s \neq \mathcal{Y}_t$), or different conditional probability distributions ($P(Y_s|X_s) \neq P(Y_t|X_t)$).
\vspace{-2mm}
\subsubsection{Transfer Learning}
Given a source domain, $\mathcal{D}_s$ and corresponding learning task $\mathcal{T}_s$, a target domain $\mathcal{D}_t$ and learning task $\mathcal{T}_t$, the objective of transfer learning is to learn the conditional probability distribution $P(Y_t|X_t)$ in $\mathcal{D}_t$ with the information gained from $\mathcal{D}_s$ and $\mathcal{T}_s$, where either $\mathcal{D}_s \neq \mathcal{D}_t$ or $\mathcal{T}_s \neq \mathcal{T}_t$ \cite{ruder2017}.
\vspace{-1mm}
\subsubsection{Transfer Learning Problems Categorization}
In the literature, TL problems are categorized in many different ways. Traditionally, TL problems are categorized based on the similarity between domains and on the availability of labeled and unlabeled data \cite{pan2010} into Inductive TL, Transductive TL and Unsupervised TL. When labeled data is available only in source domain, it is called Transductive TL and when labeled data is available in both source and target domain, it is called Inductive TL. When there is no labeled data in both source and target domain, it would be called Unsupervised TL. However, in the recent years, a flexible taxonomy \cite{weiss2016} has emerged, which is based on the domain similarity irrespective of the availability of labeled and unlabeled data, as Homogeneous TL and Heterogeneous TL. In Homogeneous TL, the source and the target domain both have the same feature space ($\mathcal{X}_s = \mathcal{X}_t$ and $\mathcal{Y}_s = \mathcal{Y}_t$), whereas in Heterogeneous TL, source and target domains have different feature spaces($\mathcal{X}_s \neq \mathcal{X}_t$ and/or $\mathcal{Y}_s \neq \mathcal{Y}_t$).
\label{subsec:tlrenewable}
 In surveys \cite{pan2010} and \cite{weiss2016}, the authors divide the general TL methods based on the relationship between the source and the target domain and give a summary of the literature on TL including details of classic TL methods. 
 \vspace{-1mm}
 \subsubsection{Transfer Learning for Organic Computing Systems} 
 Organic Computing (OC) systems aim to adapt themselves to situations that have not been foreseen by the designers. To enable this \textit{self-adaptation} in OC systems, \textit{self-learning} capability is desired. ML helps in achieving the self-learning properties for a system to adapt and learn from different conditions and environments. The relation to ML for an OC system, and a clarification of different terms in the context of these research fields is given in \cite{mbieshaar2017}.
 When a process observed or controlled by an OC system changes or when the OC system is faced with a new situation, TL plays an important role, to enable the system to adapt to changes quickly based on the previously acquired knowledge. TL also helps in making a system more autonomous and to achieve the self-learning capability.

 \subsection{Transfer Learning in Renewable Energy Systems}
 \label{subsec:tl_def}
 The TL literature on renewable energy applications is still in its early stages. In \cite{schreiber2019}, the author describes a TL framework for providing power forecasts throughout the life-cycle of wind farms and provides a taxonomy for this application, whereas in \cite{hu2016}, authors use Multi-Task Learning (MTL) autoencoders for short term wind speed predictions using TL. In \cite{svogt2020}, authors describe anomaly detection application using autoencoders for energy systems. In \cite{gensler2016},  \cite{gensler2018} the authors present the coopetitive soft-gating ensemble (CSGE) method, which weighs the ensemble
members' predictions in a hierarchical two-stage ensemble prediction system.

\section{Research Proposal}
\label{sec:proposal}
This section provides details of the research proposal in different aspects as subsections. In subsection \ref{subsec:tlframework}, we outline the TL framework using three basic What, When, and How questions in the context of TL. We also provide applicable examples from the renewable energy domain. Then in subsection \ref{subsec:tlpower}, we describe TL for regression problems based on the use case of wind power forecasting. In subsection \ref{subsec:tlanomaly}, we describe TL for anomaly detection and in subsection \ref{subsec:usecase}, we detail an use case on missing value handling for anomaly detection problems. By discussing these aspects, we evaluate the applicability of TL in renewable energy systems.

\subsection{Transfer Learning Framework for Renewable Energy Systems}
\label{subsec:tlframework}
In this section, we propose a framework for TL based on What, When, and How questions to any domain. \ref{tab:tl_framework_table} shows the exemplary application of the framework to the energy system related use cases as described previously.

\vspace{-5mm}
\begin{table}[h!]
    \caption{Transfer learning framework for renewable energy systems applications}
    \vspace{1mm}
    \centering
    \begin{tabular}{|p{0.25\textwidth}|p{0.375\textwidth}|p{0.375\textwidth}|}
    \hline
    & \textbf{Power Forecast} & \textbf{Anomaly Detection}\\
    \hline
    What to transfer?  & The most important and relevant information for $\mathcal{T}_t$ among the features of $\mathcal{D}_s$ & Selecting and picking anomaly free samples from the labeled source domain $\mathcal{D}_s$ data to train a TL model\\
     \hline
    When to transfer? & When the similarity between different wind parks based on input feature spaces $\mathcal{X}_s$, $\mathcal{X}_t$   and label spaces $\mathcal{Y}_s$, $\mathcal{Y}_t$ exceed certain threshold & When the similarity between extracted features of anomaly free data set and evaluation data set exceed certain threshold \\
    \hline
     How to transfer? & Using TL methods like parameter transfer, feature representation transfer and instance transfer between $\mathcal{T}_s$ and $\mathcal{T}_t$  & Using feature representation transfer to establish the Model of Normality (MoN) based on anomaly free samples of  $\mathcal{T}_s$. \\ 
    \hline
    \end{tabular}
    \label{tab:tl_framework_table}
\end{table}
\vspace{2mm}
\subsubsection{What to transfer?}
The answer to this question involves determining which part of the knowledge can be transferred from $\mathcal{D}_s$ to $\mathcal{D}_t$ to improve the performance of $\mathcal{T}_t$. For wind power forecasting application, we need to identify which among the $\mathcal{X}_s$ are available in $\mathcal{X}_t$ and which of them are relevant for $\mathcal{T}_t$. For anomaly detection in solar power inverter data, the first objective is to establish the MoN using anomaly free data samples \cite{aburakhia2020}. So, we need to pick the anomaly free samples from the source domain data $\mathcal{X}_s$, $\mathcal{Y}_s$, to train a TL model.

\vspace{-3mm}
\subsubsection{When to transfer?}
The answer to this question involves evaluating and deciding when TL is beneficial for a particular use case. For wind power forecasting and for anomaly detection in solar power data, a similarity evaluation between features of $\mathcal{X}_s$, $\mathcal{X}_t$ and between label spaces of $\mathcal{Y}_s$, $\mathcal{Y}_t$ is needed to determine if TL would be beneficial. The similarity between source and target tasks is then compared to a certain threshold. If it exceeds that threshold, TL can be helpful for that scenario.

The objective of TL should always be to improve the performance of the target task, but in some cases when the source and target domains are not closely related, TL leads to degradation of target task performance. This is usually termed as \textit{Negative Transfer} and such scenarios could be avoided by evaluating the answer to this question.
\vspace{-3mm}
\subsubsection{How to transfer?}
The answer to this question involves the actual methods to transfer knowledge from source to target, which is done by adapting existing source task models to the target tasks.
The general TL methods are instance transfer, parameter transfer, and feature representation transfer \cite{weiss2016}. These methods are applicable to all TL scenarios.

\begin{itemize}
\item \textbf{Parameter transfer:} This method of transfer tries to find shared parameters, which are used to transfer knowledge from $\mathcal{D}_s$  to $\mathcal{D}_t$. This approach assumes that models of related tasks share some parameters or a prior distribution of hyper parameters. In general, there are two ways to share weights among neural network models, \textit{Hard-Parameter Sharing} and \textit{Soft-Parameter Sharing}. In hard-parameter sharing, all hidden layers of neural networks are shared between source task and target task while using different output layers\cite{ruderMTL2017,ruderbingel2017}. In soft-parameter sharing, a deep neural network is trained on source domain data and the model's weights are fine-tuned using the target data. Parameter transfer is also applicable in MTL methods, where multiple tasks are learned at the same time. An example of MTL application in renewable energy systems is when an ML model is trained to predict the power generation of multiple wind farms at once.

\item \textbf{Instance transfer:} In an ideal scenario of TL, knowledge from the source domain can be directly reused for the target domain, but in most cases the knowledge can not be used directly, instead some instances of source domain can be reused along with target domain data to improve results. The idea behind instance-based TL methods is to reweigh the source instances so as to lower the marginal distribution differences \cite{chawla2002,yao2010}. With this reweighing, the source data points ($\mathcal{X}_s$, $\mathcal{Y}_s$) that are most relevant to the target domain are used along with target task data $\mathcal{Y}_t$ to train a model. 
\vspace{2mm}
\item \textbf{Feature representation transfer:} This method of TL aims to find a representation of input features by reducing the gap between source and target feature spaces and can be used for both homogeneous and heterogeneous TL. Based on the availability of labeled data, both unsupervised and supervised methods can be applied for feature representation transfers. An autoencoder \cite{hu2016}, which is an unsupervised learning strategy, is popular for this type of TL. It tries to encode or compress the input features to lower dimensions and then decode the original features from the compressed representation. This process of encoding and decoding enables an autoencoder to learn the common feature representation between source and target domains. An autoencoder can also be used for MTL tasks, where it learns a generic source representation from all tasks, which is then fine tuned to a specific new target task.
\end{itemize}
\vspace{-5mm}
\subsection{Transfer Learning for Power Forecast}
\vspace{-1mm}
\label{subsec:tlpower}

The power prediction for different wind power plants require, a separate model for each plant, which uses local numerical weather data to predict the power under the characteristic conditions of each plant. This makes it suitable for TL. 
In this section, we evaluate the application of TL for power forecasting with a focus on wind power plants.
\vspace{-3mm}

\begin{figure}[ht]
\centering
  \centering
  \includegraphics[width=\linewidth ]{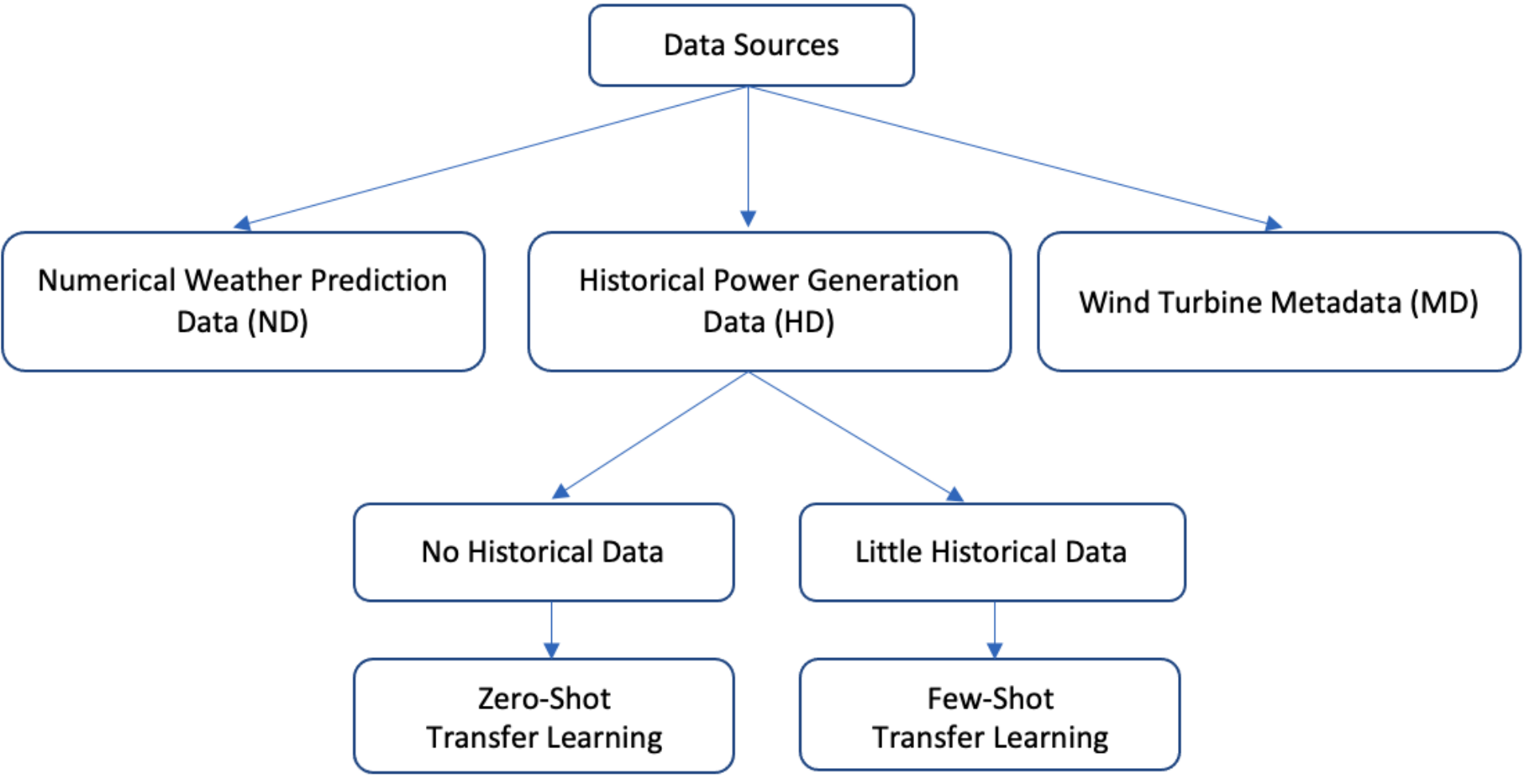}
  \caption{Input data and transfer learning methods for wind power forecasting}
  \label{fig:wind_power_forecast}
\end{figure}

% \subsubsection{Use Case - Power forecast for wind power plants}:
\vspace{-5mm}
\subsubsection{Challenges:}
There are primarily two challenges that motivate the research in this direction and are listed below:
\vspace{-2mm}
\begin{itemize}
    \item In TL scenarios, the amount of available target data usually varies.
    \item The research in TL for regression tasks, especially in renewable energy applications, is still not mature.
\end{itemize}
\vspace{-3mm}
\subsubsection{Approach:}
Generally, there are three different types of data sources available for wind power plants generally and there are two different scenarios that can arise depending on the available amount of target task data. This section deals with handling both of these scenarios using TL.

\subsubsection{Data Sources:}
\vspace{-1mm}
\begin{itemize}
\item \textit{\textbf{Numerical Weather Prediction Data (ND)}} -  This is generated by weather models, which is comprised of measurements like wind speed, wind direction, air density, wind pressure, temperature, relative humidity. There could be multiple weather models providing data, but the weather features provided would be almost similar for all of them.
\item \textit{\textbf{Historical Power Generation Data (HD)}} - Historical wind power generation data is only available for existing wind farms, while the ND and MD will be mostly available for all plants, be it newly set up or existing plants.This data will not be available for newly setup wind farms, as there won't be any historical data recorded for them. The historical data contains information regarding the power generated by a particular wind farm over a period of time.
\item \textit{\textbf{Wind Turbine Metadata (MD)}} - This data usually comprises physical characteristics of wind turbines like rotor diameter, make and model of turbine, hub height, total height, rated capacity, and year of installation.
This data set is generally available for all kinds of wind farms starting from newly installed wind farms to long existing wind farms, as it is provided by the manufacturer of the wind turbines. However, there could also be scenarios where this data is lost or not recorded for very old wind farms.

\end{itemize}
\subsubsection{TL Scenarios:}
\vspace{-1mm}
The pictorial representation of these scenarios is given in Fig.\ref{fig:wind_power_forecast}
\begin{itemize}
    \item \textit{\textbf{When there is no historical power data (no HD): }} In this scenario, the target wind farm is newly set up and there is no historical data available for power forecasting. In this case, we find a similar wind farm as a source task, based on a similarity measure. Then we use the source task model to do TL and adapt to a new plant using MD and WD. We propose \textit{Zero-Shot Transfer Learning} (ZSTL)  methods for this scenario.
    
    Zero-shot learning is interpreted as unsupervised transductive TL \cite{jreis2018}. It is transductive, as typically there is no labelled historical data available for the target task. In this setting, the metadata information (MD) is used to determine similarity between source and target task and decide which source task is to be considered for transferring the knowledge to the target task. ZSTL becomes more challenging if the meta information of tasks is also unavailable. There is a need for research here, as most of zero-shot learning (ZSL) literature is on classification problems related to computer vision and NLP. Here for renewable energy systems, specially for power forecast applications, we are interested in solving for regression problems.

    \vspace{1mm}
    \item \textit{\textbf{When there is little amount of historical power data (little HD): }} In this scenario the target wind farm has been set up for a very short duration and we have some power generation values recorded for this plant. Here, again, we find the most similar wind farm as a source task and during TL, in addition to MD and WD, we also use little or the small amount of HD available, for adapting it to the target plant. We propose \textit{Few-Shot Transfer Learning} (FSTL) methods for this kind of scenario.

    Few-shot learning refers to scenarios where are only few labeled samples are available for model training. Similar to ZSTL, this is also a less explored area of research for regression problems. However, it has been well researched for classification problems \cite{sun2019}. Domain adaptation is one feasible solution for FSTL when the target domain data has same features as the source domain data.

\vspace{2mm}
\vspace{1mm}

\end{itemize}
\vspace{-2mm}

\vspace{-5mm}
\subsection{Transfer Learning for Anomaly Detection}
\label{subsec:tlanomaly}
Anomaly detection (AD) usually deals with learning the patterns of anomalies in labeled training data and predicting upcoming failures or anomalies based on the learned patterns. Anomalies are infrequent, irregular, unexpected, and limited. The occurrence of anomalies is rare and getting labels for these is difficult. TL serves helpful in this case, as we use the label information available at plant A and transfer them to construct an AD algorithm for plant B, for which we do not have any labeled data.

In this section, we first outline the general process of anomaly detection using transfer learning and then describe current progress in handling missing data for AD problems.

\begin{figure}[ht]
\centering
\begin{minipage}{\textwidth}
  \centering
  \includegraphics[width=\linewidth]{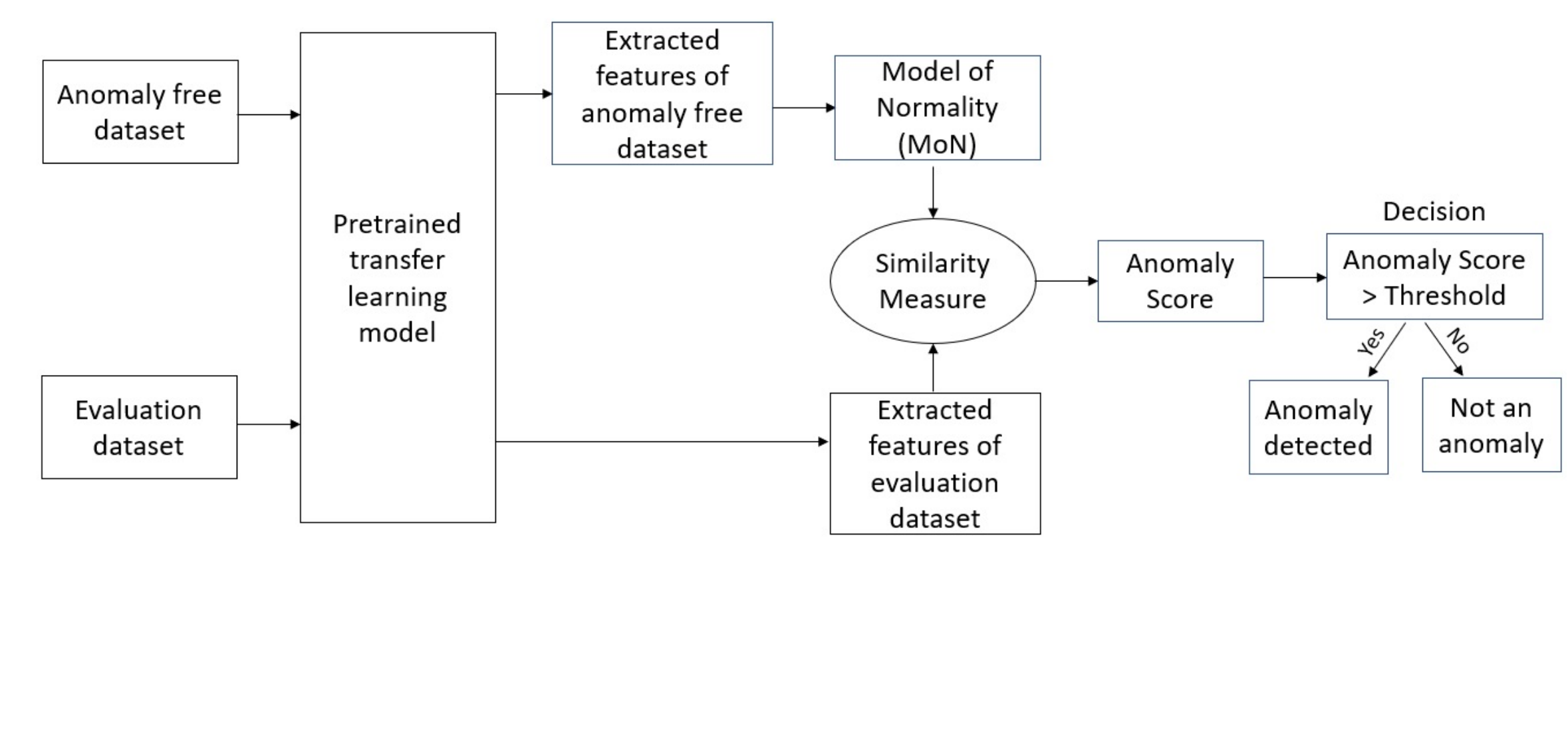}
  \captionof{figure}{Anomaly detection using transfer learning \cite{aburakhia2020}}
  \label{fig:anomaly_detection}
\end{minipage}
\end{figure}

\subsubsection{Anomaly Detection Process}
The anomaly detection process flowchart shown in Fig \ref{fig:anomaly_detection}, is proposed initially in \cite{aburakhia2020} for anomaly detection in images data. Here, we are applying it to detect anomalies in solar inverter sensor data. As a first step, we need to pick normal (anomaly-free) samples from the labeled data of normal and anomalous samples available for the source task, and feed only normal samples into the pretrained TL model. This is done to learn the pattern of normality from these anomaly free samples. The pre-trained TL model, usually a deep neural network, is trained on all inverter data and acts as a feature extractor. Features extracted from the anomaly free samples are then used to create a MoN that learns normality pattern. For any evaluation sample, which is new inverter data, the features are extracted by the TL model and compared to the MoN. Using a similarity measure, anomaly score is calculated based on which anomaly will be detected if anomaly score is above certain decision threshold.

We aim to do experiments using different types of pretrained TL models that can be used for feature extraction and compare the performances.

\subsection{Use Case on Missing Sensor Data for Anomaly Detection }
\label{subsec:usecase}
\vspace{1mm}In other ML problems generally, we either ignore missing data or impute it, but for AD applications, we can not do that as missing data itself serves as an indicator to detect anomalies or failures. In some of the AD problems, the failures in sensor data will be denoted by missing values. We describe below our first approach for dealing with missing data along with the current results. The work on this use case is still in progress.
\vspace{-2mm}
\subsubsection{Challenges} There are primarily three challenges that motivate the research in this direction, which are listed as follows. These challenges are observed in the field when dealing with solar power inverters data that has labeled failure times for inverters.
\begin{itemize}
    \item How to handle missing time series sensor data for an AD use case? 
    \item How to deal with variable input feature space, when some sensors exhibit missing data?
    \item How many of the available sensors are related to each other? 
\end{itemize}
\vspace{-2mm}
\subsubsection{Approach:} All the above-mentioned challenges are interlinked and can be solved in some form with the method below.
\vspace{-2mm}
\subsubsection{Data Source:} We generated a synthetic data set of 1000 samples with pairs of inputs $\mathcal{X}_i$ and target $\mathcal{Y}$. For each sample, ten input features $\mathcal{X}_1$ to $\mathcal{X}_{10}$ are generated from two latent features $\mathcal{Z}_1$, $\mathcal{Z}_2$, such that they belong to three groups of features. The latent features $\mathcal{Z}_1$ and $\mathcal{Z}_2$ are related to target $\mathcal{Y}$ according to Equation 1

\begin{eqnarray}
\mathcal{Y} = \sin{\frac{\pi}{4}}(\mathcal{Z}_1 + \mathcal{Z}_2) + \cos{0.7\pi(\mathcal{Z}_1 - \mathcal{Z}_2)}
\end{eqnarray}

The latent features $\mathcal{Z}_1$, $\mathcal{Z}_2$ are i.i.d sampled from a standard normal distribution $\mathcal{Z} \sim \mathcal{N}(0,1)$. 
The first group of features is formed according to eq (\ref{firstgroupeqn}), while the second group of features is formed according to eq (\ref{secondgroupeqn}) where $\epsilon \hspace{1mm} \sim \hspace{1mm} \mathcal{N}(0,0.1)$. Furthermore, the third group is formed by $\mathcal{X}_i \sim \mathcal{N}(0,1) \hspace{1mm} \forall \hspace{1mm} i \in \{7,8,9,10\}$.

\begin{eqnarray}
\mathcal{X}_i = \mathcal{Z}_1 + \epsilon \hspace{1mm}  \forall i \in \{1,2,3\}
\label{firstgroupeqn}
\end{eqnarray} 
\vspace{-2mm}
\begin{eqnarray}
\mathcal{X}_i = \mathcal{Z}_2 + \epsilon \hspace{1mm}  \forall i \in \{4,5,6\}
\label{secondgroupeqn}
\end{eqnarray}

To create the missing values in this data, during training, features are removed randomly with a probability of 0.2 from the binomial distribution.

\subsubsection{Feature Embedding Experiment:} 
\begin{figure}[ht]
\centering
% \begin{minipage}{.48\textwidth}
  \centering
  \includegraphics[height=0.4\textheight]{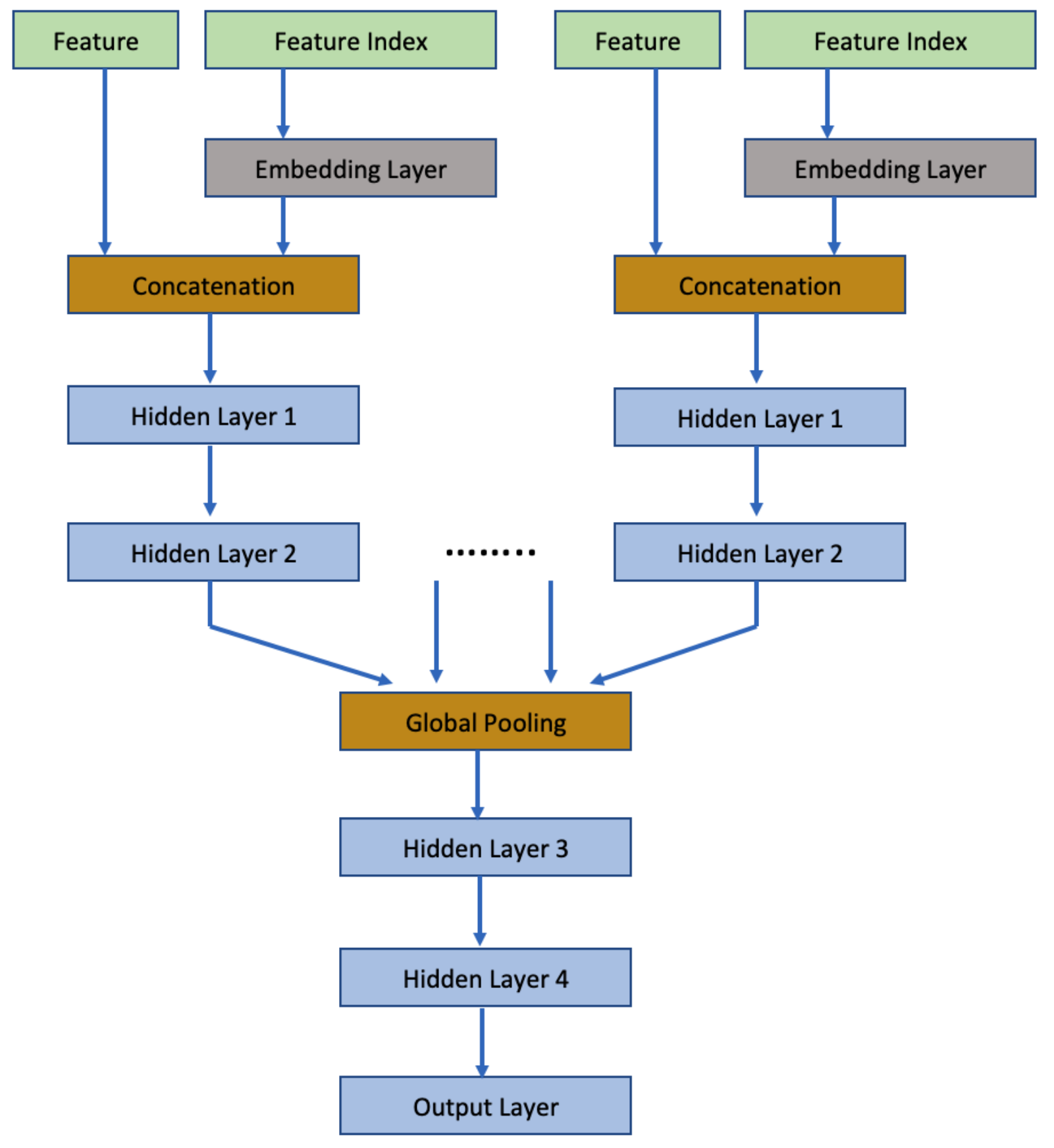}
  \caption{Neural network architecture for feature embedding method}
  \label{fig:nnarch}
% \end{minipage}%
\end{figure}

This approach deals partially with the problem of missing input features and partially with the problem of training with variable input lengths of features. Instead of adopting a fixed input vector length, this approach tries to generalize the concept of input features, by describing their properties and relevance for the task at hand with individual feature embeddings. The architecture of the neural network for this approach is given in Fig. \ref{fig:nnarch}. 

Here, the feature index, $i$ in $\mathcal{X}_i$ is an input to embedding layer and then each feature is represented with its value concatenated with the output of the embedding layer, which is an embedding vector. These concatenated features are now processed in two hidden layers. Up to this point, the features are processed individually in per-feature neural networks which are termed as \textbf{feature encoders}. The outputs of these feature encoders corresponding to all available features are now averaged in the global pooling layer similar to global averaging in CNN networks to get a representation, independent of the current number of input features. The resulting averaged representation is then further processed to compute the final model output, with respect to the task at hand. The embeddings are learned along with other parameters of the neural network and serve as task-specific latent representations of input feature space. 

\begin{figure}[h]
% \begin{minipage}{.48\textwidth}
  \centering
  \includegraphics[width=0.8\linewidth]{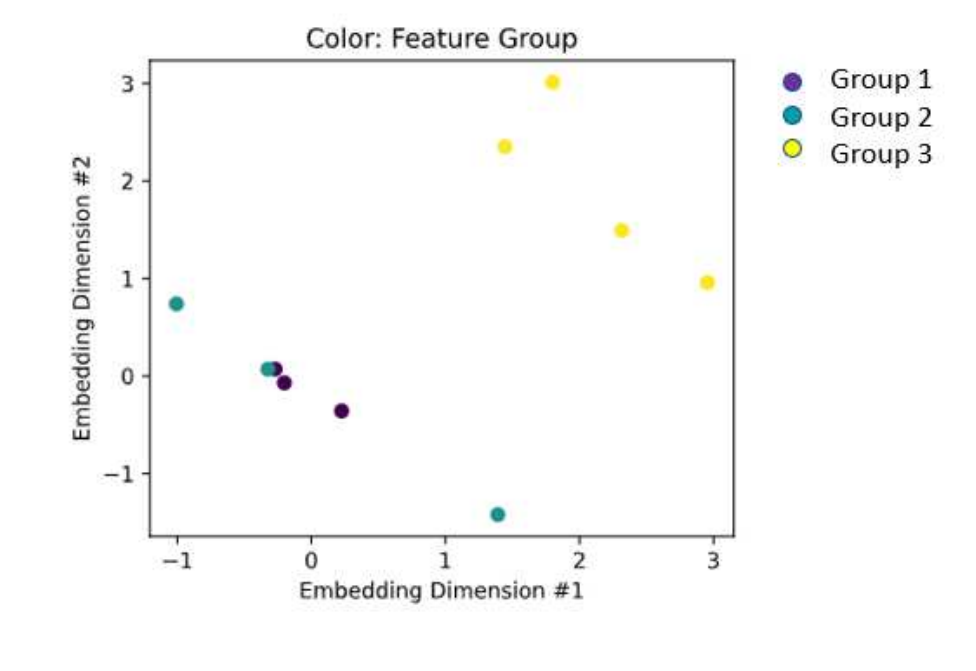}
  \caption{Visualisation of 2-D feature embedding}
  \label{fig:featuregroups}
% \end{minipage}
\vspace{-5mm}
\end{figure}

The neural network training with this method, on synthetic data generated with the above mentioned equations, derived the two-dimensional embeddings. The features can now be easily interpreted with Fig. \ref{fig:featuregroups}. In this figure, we can observe that the related features are located closer to each other, creating groups of features in the latent space. Here, we can see that groups of features are clustered and that these clusters are appropriate, as we already know the features of Group1 are related, since they are generated from same latent feature $\mathcal{Z}_1$. The observations for Group2 are similar. The plot helps in understanding which features are contributing similar information to the model. We are also exploring other ways of solving the challenges mentioned above using embeddings with different neural network architectures.

The feature embedding method provides the expected results on the synthetic data set. The next step is to evaluate this approach on real-world renewable energy data sets to further assess the feature embedding methodology.

\vspace{-3mm}

\section{Progress Description and Summary}
\vspace{-1mm}
\label{sec:progress}
In this section, we summarize our preliminary work towards the research proposal of section \ref{sec:proposal}. Currently, different TL methods for wind power forecasting and solar inverter data anomaly detection are being tested on real world datasets. The broader aspects of future work are outlined below.  
\begin{itemize}
    \item\textbf{Evaluating Similarity Measures} for deciding which source task is most similar to a certain target task
    \item\textbf{Task Embeddings} for multi-task learning methods of TL
    \item\textbf{Zero-shot and Few-shot TL} methods for scenarios of transfer with low amounts of data for the target task
\end{itemize}

Conclusively, this article proposes a TL framework and methodologies for renewable energy system power forecasting and anomaly detection. We also propose a feature embedding approach, which can potentially solve problems with regards to missing data. This proposal is an initial step towards exploring the opportunities of applying TL in renewable energy systems. Further, the TL methods in this proposal are helpful for building OC systems' self-learning capability. This cross-domain applicability of TL methods further motivates the research in this area.

\section*{Acknowledgments}
This work is supervised by Prof. Dr. Bernhard Sick and Stephan Vogt and supported by the Digital-Twin-Solar (03EI6024E) project, funded by the BMWi (German Federal Ministry for Economic Affairs and Energy) and the project TRANSFER (01IS20020B), funded by the BMBF (German Federal Ministry of Education and Research).

\nocite{*}

\bibliographystyle{unsrt}
\vspace{-3mm}
\bibliography{cn:references}

\end{document}